\documentclass[11pt,en]{elegantpaper}

\usepackage{url}            
\usepackage{booktabs}       
\usepackage{multirow}
\usepackage{amsfonts}       
\usepackage{nicefrac}       
\usepackage{microtype}      
\usepackage{xcolor}         
\usepackage{makecell, rotating}
\usepackage{comment}
\usepackage{amsmath}
\usepackage{amssymb}
\usepackage{wrapfig}
\usepackage{ascmac}
\usepackage{algorithm}
\usepackage{algorithmic}

\title{DoubleCCA: Improving Foundation Model Group Robustness with Random Sentence Embeddings}
\author{Hong Liu\footnote{E-mail: lynnliu.xmu@gmail.com} \\ Osaka University \and Yitong Lu \\ Sichuan University}
\date{}

\begin{document}

\maketitle

\begin{abstract}
    This paper presents a novel method to improve the robustness of foundation models to group-based biases. 
    We propose a simple yet effective method, called DoubleCCA, 
    that leverages random sentences and Canonical Correlation Analysis (CCA) to enrich the text embeddings of the foundation model.
    First, we generate various random sentences that augment the original prompts, which extends the original prompts with random words or character sequences.
    Second, we use an additional sentence embedding model to generate different text embeddings with respect to these random sentences.
    We then use CCA double twice to align the representations and reconstruct them back to the original representation space. 
    We demonstrate the effectiveness of our method on a variety of tasks and datasets, showing that it outperforms existing methods in terms of both performance and robustness. 
    Our method is simple to implement and can be easily integrated into existing models, making it a practical solution for improving the robustness of foundation models to group-based biases.\par
\end{abstract}


\section{Introduction}\label{sec:intro}

Contrastive language-image pretraining (CLIP) and its variants \cite{radford2021learning,zhai2023sigmoid,desai2023hyperbolic} are the widely used vision-language models (VLMs).
They usually train models on large-scale datasets with a large number of image-text pairs, such as LIAON-400M \cite{schuhmann2021laion}.
Recent works have shown impressive zero-shot generalization on a wide range of tasks, such as medical image classification \cite{wang2022medclip}, object detection \cite{ramaswamy2024geode} and semantic segmentation \cite{sun2024clip,licascade}.

Recent works \cite{menon2022visual,roth2023waffling,an2024perceptionclip} show that current VLMs lack systematic investigation of the prompts they used. 
Therefore, they propose to modify the prompts to improve the model's performance, especially the ability of domain generalization.
Despite their success of the remarkable zero-shot capability, the foundation models is still sensitive to the group-based biases, which are the attributes that are correlated with the ground-truth labels but are not directly related to the classification task.

\begin{figure}
    \centering
    \includegraphics[width=0.47\textwidth]{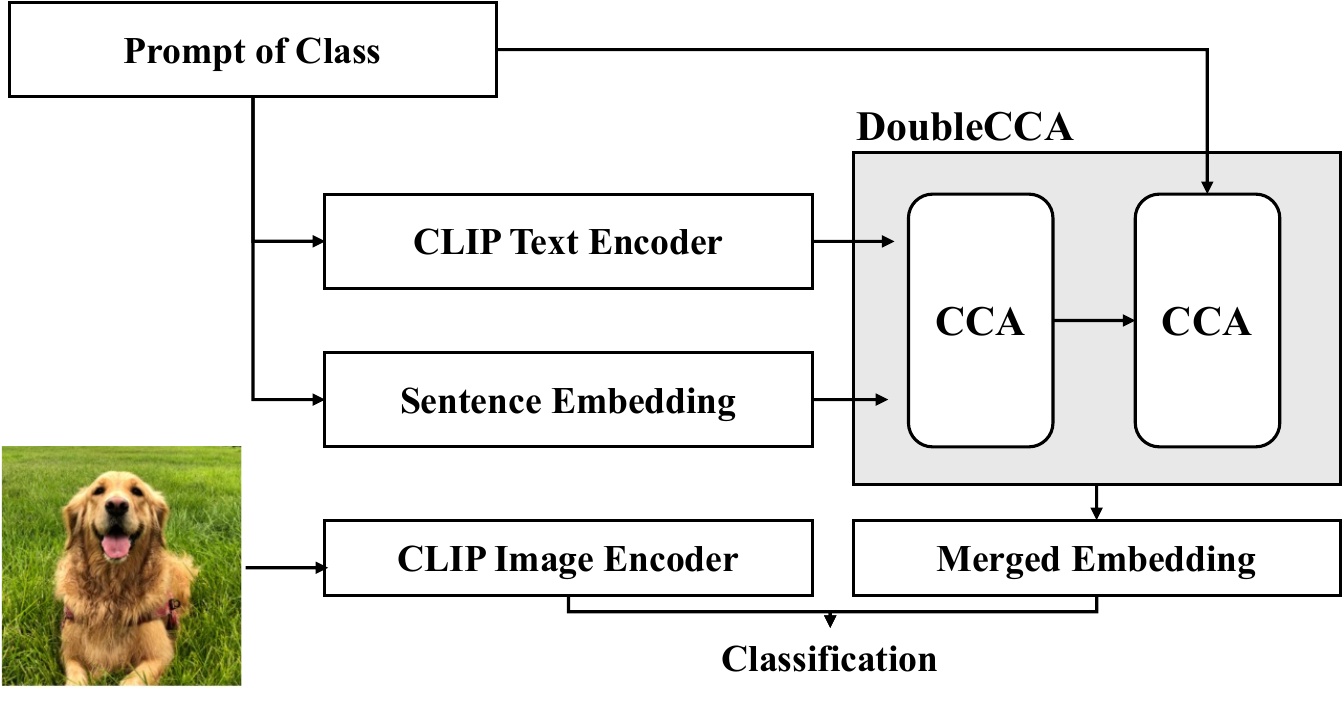}
    \caption{The pipeline of our proposed DoubleCCA. We leverage random words to augment semantic descriptions and introduce an additional sentence embedding model to complement the semantic limitations of the original VLM text encoder. We use classical CCA technique double twice to merge different semantic information, which helps to improve the group robustness of the CLIP model.
    }
    \label{fig:teaser}
\end{figure}

A robust classifier should evade the influence of irrelevant features present in the image or text data. Therefore, it is necessary to make the classifier invariant to the group attributes.
For example, in the Waterbirds dataset, the background of the image is a group attribute that is correlated with the ground-truth labels, but it is not directly related to the classification.
To improve group robustness, there are a lot of works that focus on model debiasing \cite{zhang2022contrastive,kumar2022finetuning,kirichenko2023last,chuang2023debiasing,dehdashtian2024fairerclip,you2024calibrating,gao2024clip,pmlr-v235-phan24b,yang2024debiasing}. 
Most of these works aim to add a simple adapter architecture to the end of the CLIP model, and then update the parameters of the adapter on a dataset with target labels and group attributes.
Although these methods have shown some improvements in terms of group robustness, they still have some limitations.

First, the performance of the model is highly dependent on the dataset used for training the newly added adapter architecture. 
This will hinder the generalization ability of the model to other datasets efficiently, which means that they cannot improve the zero-shot ability of the CLIP model. 
Second, some of the works \cite{chuang2023debiasing,yang2024debiasing} employ prompt tuning techniques to improve the performance and fairness of the model. 
Yang et al. \cite{yang2024debiasing} utilize LLM to synthesize a balanced text dataset, and then use the prompt tuning to improve the model's performance.
This method also relies on the prior knowledge of the dataset or large language model, and it is difficult to quickly generalize to other dataset or requires additional API costs.
Therefore, current debiasing methods still have some limitations in terms of generalization ability and efficiency of the model.

Therefore, we ask the following question: \textit{ How can we improve the group robustness of the foundation model without relying on prior knowledge of the dataset?}
To answer this question, we propose a novel method, called DoubleCCA, to improve the robustness of foundation models to group-based biases.

Inspired by \cite{roth2023waffling}, we use random words or character sequences to augment the original prompts, thus enriching the representation space of text embeddings.
We call these generated sentences as the random sentences.
Then, we use the CLIP text encoder and the extra sentence embedding model (such as the Hierarchy Transformer Encoder \cite{he2024language}) to generate the different types of text embeddings with respect to these random sentences.
We use Canonical Correlation Analysis (CCA) to map the representations of different two embedding features into a common feature space.
We then utilize CCA again to merge the embedded representations and reconstruct them back to the original representation space to align with CLIP's visual features.
The proposed DoubleCCA method can be integrated into most existing VLMs, which can improve the group robustness of the foundation model without relying on the prior knowledge of the dataset.
The pipeline is shown in Figure \ref{fig:teaser}, and
our contribution can be summarized as follows.
\begin{itemize}
    \item We propose a novel method, called DoubleCCA, to improve the group robustness of foundation models to group-based biases.
    \item We leverage random words to augment text descriptions and introduce an additional sentence embedding model to complement the semantic limitations of the original CLIP text encoder through the CCA technique.
    \item We show the effectiveness of our method on a variety of datasets, showing that it outperforms existing methods in terms of both group robustness and domain generalization.
\end{itemize}

\section{Preliminaries}\label{sec:prelim}

This section will introduce the necessary background knowledge for our method, including the CLIP foundation model and Canonical Correlation Analysis (CCA).

\noindent\textbf{CLIP model.}
CLIP model \cite{radford2021learning} is a vision-language foundation model that consists of two parts: a vision encoder and a text encoder.
The vision encoder $\Phi_v: \mathbb{R}^{d_v}\rightarrow \mathbb{R}^d$ and the text encoder vision encoder $\Phi_t: \mathbb{R}^{d_t}\rightarrow \mathbb{R}^d$ are deep models that map the input image and text to a $d$-dimensional embedding space, respectively.
Given a batch of image-text pairs $(I, T)$, the model is trained to minimize symmetric contrastive loss \cite{radford2021learning}, which aligns the image-text embedding pairs in the representation space $\mathbb{R}^d$.

Once the model is trained, we can directly use image and text encoder to align images with text descriptions. Thus, a zero-shot image classifier can be built by comparing the similarity between the image embedding $\Phi_v(I)$ and the text embedding $\Phi_t(T)$.
The typical method is to combine the name of the class $k$ with the predefined template to obtain the text description $t_k$.
For example, the class of zebra can be integrated into the prompt template ``a photo of a $\langle$class name$\rangle$'' to yield the description ``a photo of a zebra''.
Thus, we can compute the logits for each class by the cosine similarity between the image embedding and the text embedding, and the class with the highest score is the predicted class.

\noindent\textbf{Canonical Correlation Analysis (CCA).}
Canonical Correlation Analysis (CCA) is a statistical method that finds the transformation that maximizes the correlation between two feature sets from different models. 
Let $X_A \in \mathbb{R}^{n \times d_A}$ and $X_B \in \mathbb{R}^{n \times d_B}$ be the data matrices, where $n$ is the number of samples, and $d_A$ and $d_B$ are the dimensions of the feature vectors.
CCA finds the transformation matrices $W_A$ and $W_B$ that maximize the correlation between the transformed features $Z_A = X_AW_A$ and $Z_B = X_BW_B$ in a common feature space. 

We further define $S^{XX} = X_A^TX_A$ and $S^{YY} = X_B^TX_B$ as the covariance matrices of $X_A$ and $X_B$, and $S^{XY} = X_A^TX_B$ as the cross-covariance matrix.
Therefore, the formulation of CCA can be written as follows:
\begin{equation}\label{eq:cca}
    \begin{aligned}
        \max_{W_A, W_B} & \quad \text{corr}(Z_A, Z_B) = W_A^TS^{XY}W_B \\
        \text{s.t.} & \quad W_A^TS^{XX}W_A = I, \quad W_B^TS^{YY}W_B = I, \\
    \end{aligned}
\end{equation}
where $\text{corr}(Z_A, Z_B)$ is the correlation between $Z_A$ and $Z_B$, and $I$ is the identity matrix.

This formulation can be solved by eigenvalue decomposition of the generalized eigenvalue problem:
\begin{align}
    U, S, V^T &= SVD\big((S^{XX})^{-1/2}\cdot S^{XY}\cdot(S^{YY})^{-1/2}\big), \nonumber \\
    W_A &= (S^{XX})^{-1/2}U, \quad W_B = (S^{YY})^{-1/2}V. \nonumber
\end{align}
In practice, we center the data before applying CCA to ensure the data has zero mean. And we use regularized CCA \cite{corrochano2005eigenproblems,horoi2024harmony} to make the computation of $W_A$ and $W_B$ more stable.

\section{Method}\label{sec:method}


In this section, we introduce our method, DoubleCCA, to improve the robustness of foundation models to group-based biases.
We first define and analyze the problem (see Sect. \ref{sec:problem}) and then present the details of our method (see Sec. \ref{sec:doublecca}).

\subsection{Problem Analysis}\label{sec:problem}

\begin{figure}[!t]
    \centering
    \includegraphics[width=0.7\textwidth]{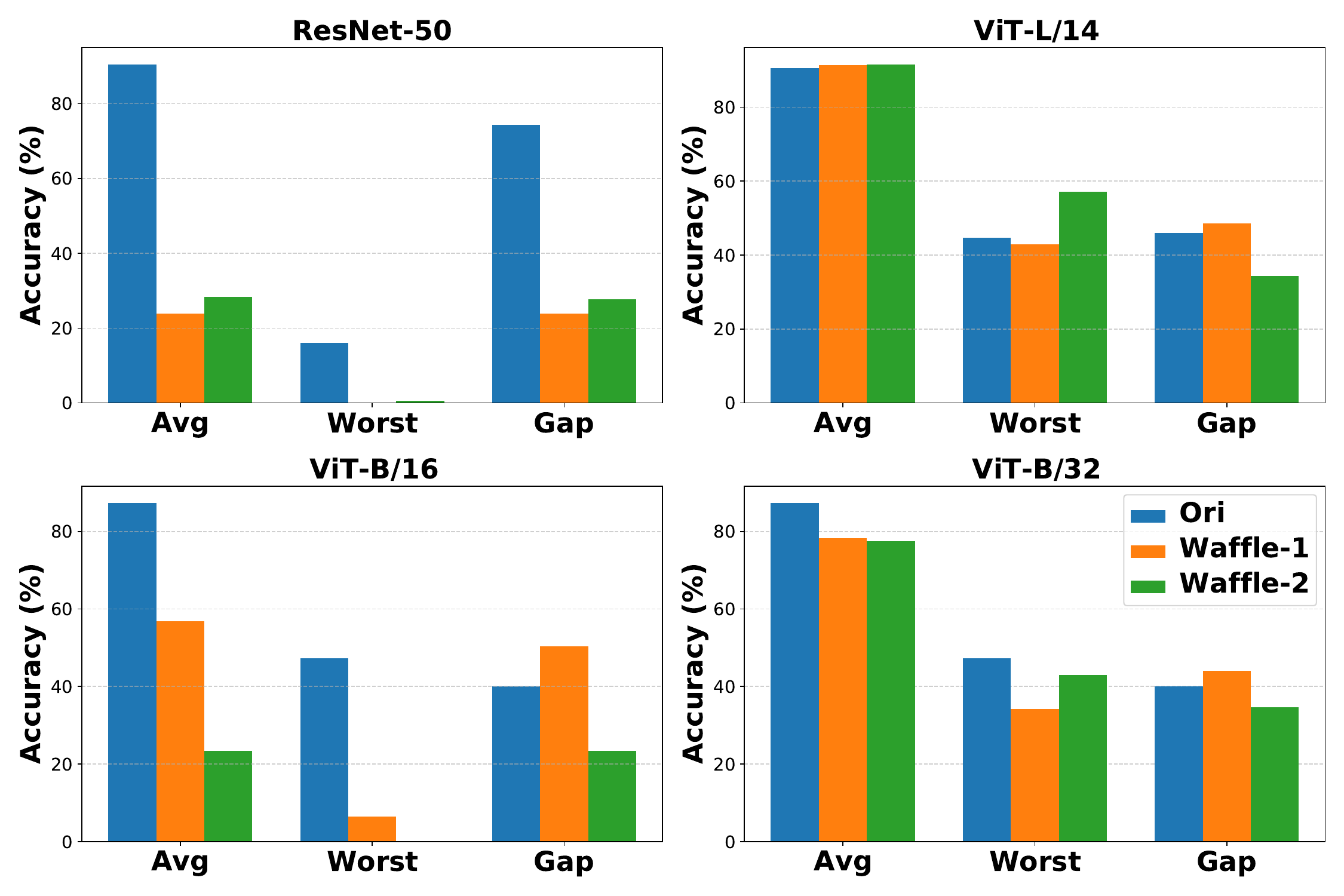}
    \caption{  
    We compare the performance of different prompts with different backbone models on the Waterbirds dataset.
    ``Ori'' denotes the original prompt of CLIP, i.e., ``a photo of a $\langle$class name$\rangle$''.
    ``Waffle-1'' denotes the combination of the original prompt and the random words, i.e., ``a photo of a $\langle$class name$\rangle$, which has $\langle$random word$\rangle$''.
    ``Waffle-2'' also denotes the combination of the original prompt and the random words, but with different template, i.e., ``a photo of a $\langle$class name$\rangle$,  $\langle$random characters$\rangle$''.
    }
    \label{fig:comparison}
\end{figure}


One interesting approach to improve CLIP's zero-shot classification is to augment the prompts with additional visual concepts from external knowledge sources.
Menon and Vondrick \cite{menon2022visual} utilize large language models (LLMs) like GPT-3 to generate class-specific descriptions for each class and incorporate them into prompts, resulting in prompts like ``a photo of a hen, which has two legs.''
But this kind of method is limited to prior knowledge of the class name and the GPT-3 generated descriptions have high degrees of ambiguity and limited visual relevance.  

Thus, Roth et al. \cite{roth2023waffling} propose a method called WaffleCLIP, which substitutes GPT-3 generated descriptors with random word or character sequences,  resulting in prompts such as ``a photo of a hen, which has jmhj, !J\#m.'' Where ``jmhj, !J\#m'' is the random character sequences. 
Based on WaffleCLIP, we simply study the effect of this method on the group robustness of the CLIP model.
We conduct four toy experiments on the Waterbirds dataset \cite{Sagawa2020Distributionally} with four different backbone models, i.e., ResNet-50, ViT-B/32, ViT-B/16, and ViT-L/14,
We compare the results of vanilla CLIP with the original prompt, WaffleCLIP with random words, and WaffleCLIP with the random characters.
See Figure \ref{fig:comparison}.

\begin{figure}
    \centering
    \begin{tabular}{cc}
        \includegraphics[width=0.45\textwidth]{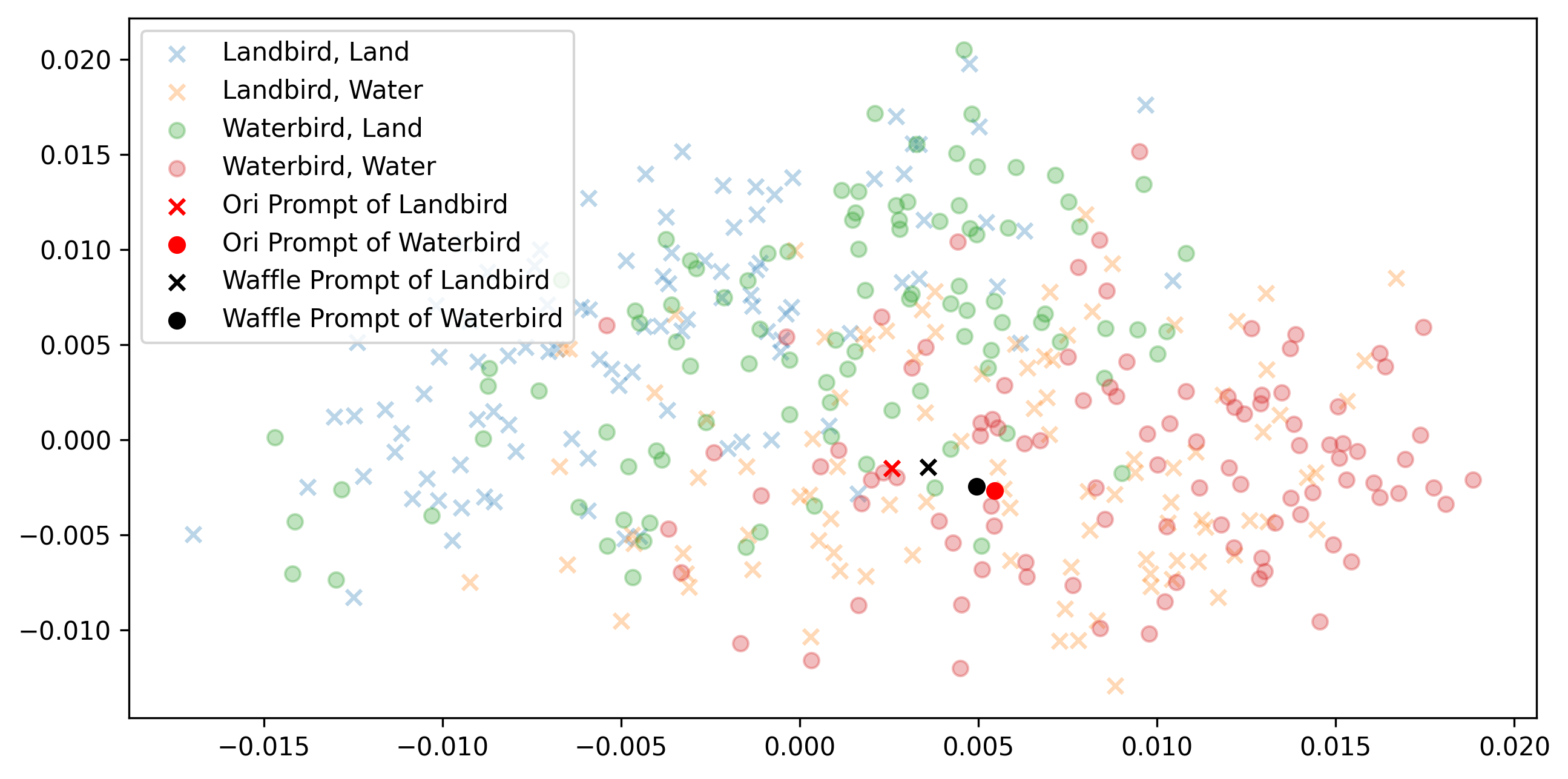} &       
        \includegraphics[width=0.45\textwidth]{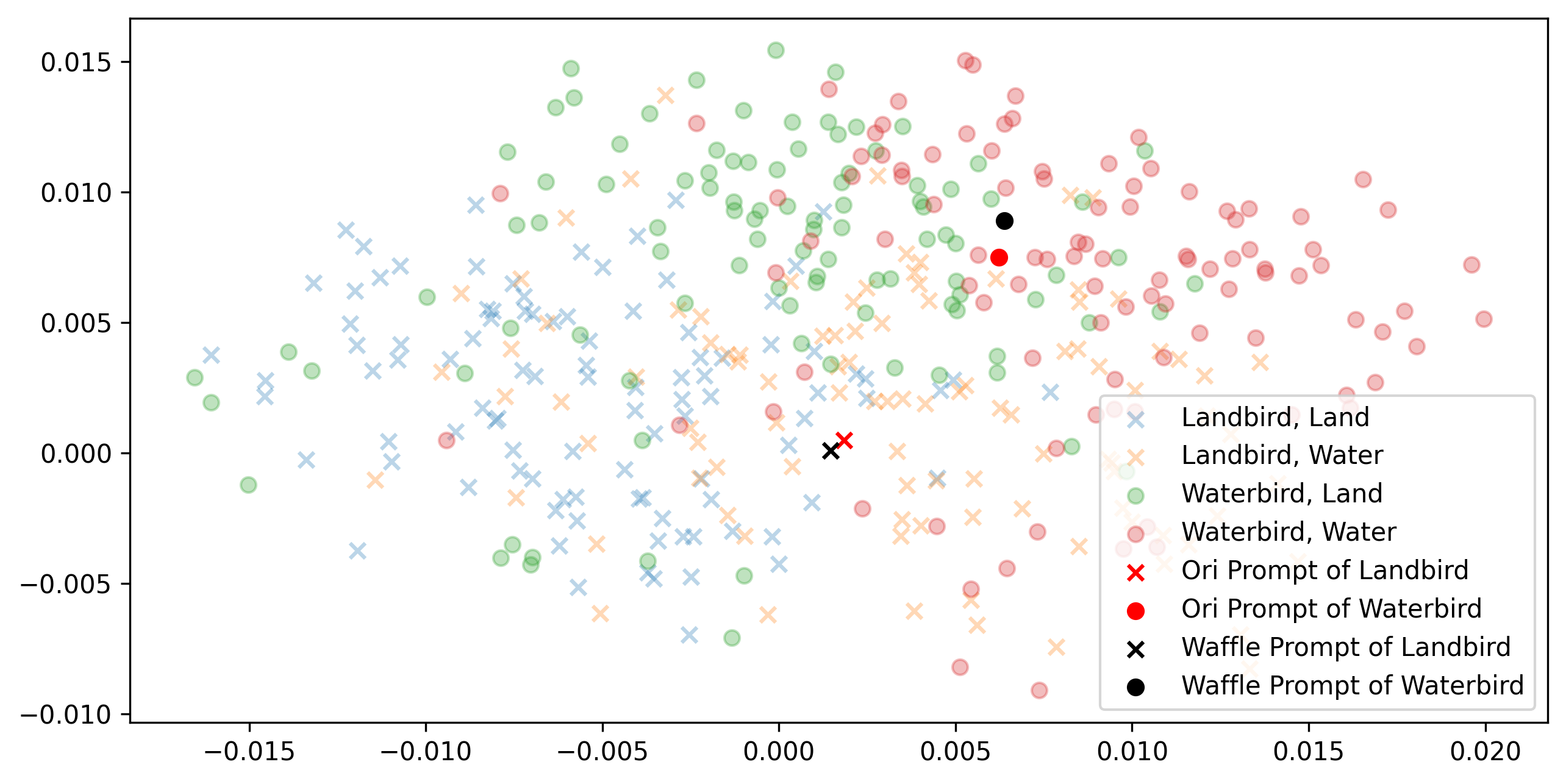} \\
         (a) RN50 &  (b) ViT-L/14 \\
    \end{tabular}
    \caption{The visualization of the image embeddings of the Waterbirds dataset. We also visualize the text embedding features extracted by the CLIP text encoder. The ``Ori prompt'' means the original prompt, i.e., ``a photo of a $\langle$class name$\rangle$''. The ``Waffle prompt'' denote the prompts with the random words and characters.
    } \label{fig:tsne}
\end{figure}

We observe that WaffleCLIP achieves better results in terms of average accuracy and worst group robustness only when using the ViT-L/14 backbone.
For the other three backbone models, its performance is worse than vanilla CLIP with the original prompt.
Moreover, when using ViT-B/16 or ResNet-50 as the backbone, WaffleCLIP's worst group robustness drops to near zero, which is substantially lower than that of the original prompt.
Thus, WaffleCLIP is not stable and is highly dependent on the vision backbone model.
To analyze the reason behind this phenomenon, we visualize the representations of the image embeddings and the text embeddings of the Waterbirds dataset using t-SNE \cite{maaten2008visualizing}. See Figure \ref{fig:tsne}.

First, for visual features, we observe that the ViT-L/14 model has a clear separation between the different groups, while the ResNet50 model has a large overlap between the different groups.
Second, for text features, we find that:
adding random words or characters to the prompt can pull the representations of different classes closer together, which is not beneficial for the group robustness. However, the different phenomena are observed in the ViT-L/14 model, where the text features will push away from each other.
Science we add the same random words or characters to the prompt, the different backbone models will generate different text embeddings, which will lead to different results.

The fundamental limitation of WaffleCLIP lies in its use of meaningless semantic random words and characters. 
The text encoder generates different text embeddings for the different random words or characters, which fail to provide a stable and meaningful representation for the model to predict the class.
To address this issue, we propose to generate semantically meaningful text embeddings that can effectively enhance the model's performance and group robustness. We will detail our method in the following section.

\subsection{DoubleCCA}\label{sec:doublecca}
According to the analysis in Sect.\ref{sec:problem}, we argue that random words or characters will introduce randomness to text embeddings, leading to worse group robustness.
Therefore, our target is to generate semantically meaningful text embeddings that can effectively enhance the foundation model's performance and group robustness.

However, only using the CLIP text encoder model to generate text embeddings is not enough to achieve this goal.
In fact, this text encoder plays the same role as the sentence embedding model in natural language processing (NLP) tasks \cite{sentence-bert,he2024language}, which transforms sentences into fixed-dimensional vector representations.
Current sentence embedding models are trained on large-scale text data, and they can generate semantically meaningful text embeddings. 
Therefore, our idea is to utilize these sentence embedding models to enrich the text embeddings of the CLIP model.

There are two major challenges in this process.
First, the dimensional of the text embeddings generated by the sentence embedding model may not be the same as the text embeddings generated by the CLIP text encoder.
Second, it is difficult to merge these new generated sentence embeddings into the CLIP model. 

To address these challenges, we propose a novel method, called DoubleCCA, which utilizes canonical correlation analysis (CCA) technique twice. 
The first CCA is used to align the representations of different embeddings into a common space. 
The second CCA is to merge the aligned representations and then recover to the original embedding space. 

\subsubsection{Step 1: The First CCA.}
We first generate sentence embeddings using the sentence embedding model $\Phi_{se}$ and the CLIP text encoder $\Phi_t$. 
Let $X \in \mathbb{R}^{n \times d}$ and $X_{se} \in \mathbb{R}^{n \times d_{se}}$ be the data matrices, where $n$ is the number of classes in the dataset, $d$ and $d_{se}$ are the dimensions of the text embeddings generated by the CLIP text encoder and the sentence embedding model, respectively.

We then apply CCA (w.r.t. Eq.\ref{eq:cca}) to learn the transformation matrices $W_x$ and $W_{se}$ that embeds two features into a common space:
\begin{equation}
    Z_x = XW_x, \quad Z_{se} = X_{se}W_{se},
\end{equation}
where $Z \in \mathbb{R}^{n \times d_{cca}}$ and $Z_{se} \in \mathbb{R}^{n \times d_{cca}}$ are the aligned representations of the sentence embeddings and the CLIP text embeddings, respectively.

However, the number of class labels is usually much smaller.
For example, there are only two classes in the Waterbirds dataset.
This means that only two sentences are used for the CCA to learn the transformation matrices $W_x$ and $W_{se}$. 
We think this is not enough to learn a stable transformation matrices. (The next section will show the experimental verifications.)
To address this issue, we propose to use the random sentence embeddings to generate more sentence embeddings.
We first combine the original prompt and the random character sequences, i.e., ``a photo of a $\langle$class name$\rangle$,  $\langle$random character sequences$\rangle$''. We call this as the \textit{random sentence}.
We then generate $K$ random sentences for each class and use the sentence embedding model and the CLIP text encoder to extract the corresponding sentence embedding features, i.e., $F_{rse}$ and $F_r$ respectively.
We replace $X_{se}$ with $F_{rse}$ and $X$ with $F_r$ to apply CCA to learn the transformation matrices $W_x$ and $W_{se}$.

\subsubsection{Step 2: The Second CCA.}
Let us rethink the zero-shot classification process. 
The CLIP model serves as a score function, which computes the similarity between the image embedding and the text embedding, i.e., $S(I,y) = f_t^Tf_v$, where $f_v = \Phi_v(I)$ and $f_t = \Phi_t(T)$.
The CLIP method predicts the class $\hat{y} \in \mathcal{Y}$ with the highest score, i.e., $\hat{y} = \arg\max_{y\in\mathcal{Y}}S(I,y)$.

After the first CCA, we can achieve two different score functions 
\begin{align}\label{eq3}
    S_x(I,y) = {x^{(y)}}^TW_xW_x^Tf_v \\
    S_{se}(I,y) = {x^{(y)}_{se}}^TW_{se}W_x^Tf_v,\label{eq4}
\end{align}
where $x^{(y)}$ and $x^{(y)}_{se}$ are the text embeddings of the class $y$ w.r.t. the original prompts. 
In practice, we can define $\hat{W}_x = XW_xW_x^T$ and $\hat{W}_{se} = X_{se}W_{se}W_x^T$.
Eq.\ref{eq3} and Eq.\ref{eq4} can be rewritten as two fully connected layers, i.e., $\hat{Y} = \mathrm{FC}(f_v; \hat{W}_x )$ and $\hat{Y} = \mathrm{FC}(f_v;\hat{W}_{se})$, where $\hat{Y}$ is the predicted logit feautre w.r.t. the input image feature.

In fact, combining the predictions of these two models through ensemble learning is a good way to improve accuracy. 
Referring to \cite{horoi2024harmony}, we employ the CCA technique to merge these two fully-connected layers into one fully-connected layer, i.e., $\hat{y} = \mathrm{FC}(f_v;W)$, where $W$ can be seen as the merged text embeddings.

The merging with CCA can be formulated as follows:
\begin{equation}
    W = \frac{1}{2}(\hat{W}_x + M \cdot \hat{W}_{se}),~M = (P_B\cdot P_A^{-1})^T,
\end{equation}
where $P_A$ and $P_B$ are the transformation matrices learned by the second CCA. 

Since we do not access to the image data, we further utilize the random sentence embedding features $F_r$ that simulate the image representations. 
Therefore, we can achieves two feature sets:
\begin{equation}
    X_A =  \hat{W}_xF_r,~~X_B =  \hat{W}_{se}F_r.
\end{equation}

\begin{algorithm}[!t]
    \caption{DoubleCCA}
    \label{alg:doublecca}
    \begin{algorithmic}[1]
        \REQUIRE Sentence embedding model $f_{se}$, CLIP model $(f_v, f_t)$, number of random sentences $K$
        \ENSURE Merged text embeddings $W$
        \STATE Generate $K$ random sentences for each class
        \STATE Extract sentence embeddings $F_{rse}$ using $\Phi_{se}$
        \STATE Extract CLIP text embeddings $F_r$ using $\Phi_t$
        \STATE Apply CCA to $X$ and $Y$ to obtain $W_x$ and $W_{se}$
        \STATE Compute $\hat{W}_x = XW_xW_x^T$, $\hat{W}_{se} = X_{se}W_{se}W_x^T$
        \STATE Generate random sentence embedding features $F_r$
        \STATE Compute $X_A = \hat{W}_xF_r$, $X_B = \hat{W}_{se}F_r$
        \STATE Apply CCA to $X_A$ and $X_B$ to obtain $P_A$ and $P_B$
        \STATE Compute $M = (P_B \cdot P_A^{-1})^T$
        \STATE Merge text embeddings: $W = \frac{1}{2}(\hat{W}_x + M \cdot \hat{W}_{se})$
        \RETURN $W$
    \end{algorithmic}
\end{algorithm}

Then, we can apply CCA to learn the transformation matrices $P_A$ and $P_B$ via maximization of the correlation between $X_A$ and $X_B$ as follows:
\begin{equation} 
    \begin{aligned}
        &\max_{P_A, P_B}  \quad \text{corr}(X_A, X_B) = P_A^TS^{AB}P_B \\
        &\text{s.t.}  \quad P_A^TS^{AA}P_A = I, \quad P_B^TS^{BB}P_B = I, \\  
        &S^{AA} = X_A^TX_A,  \quad S^{BB} = X_B^TX_B, \quad S^{AB} = X_A^TX_B.  
    \end{aligned}
\end{equation}

\subsubsection{Inference}
The overall process of DoubleCCA is summarized in Algorithm \ref{alg:doublecca}.
After DoubleCCA, we can achieve the merged text embedding matrix $W\in \mathbb{R}^{n\times d}$.
We can directly use this merged text embeddings to predict the class label of the input image, which can be formulated as follows:
\begin{equation}
    \hat{y} = \arg\max_{y\in\mathcal{Y}}S(I,y),~\text{where}~ S(I,y) = W_y\Phi_v(I),
\end{equation}
where $W_y\in \mathbb{R}^{1\times d}$ is the $y$-th row of the merged embedding matrix $W$, which is the embedding feature of the class $y$.

\section{Experiments}\label{sec:experiments}

\subsection{Experimental Setup}\label{sec:setup}

\noindent\textbf{Datasets.} 
We evaluate the group robustness of our method. We conduct experiments on two widely used datasets: Waterbirds \cite{Sagawa2020Distributionally} and CelebA \cite{liu2015deep}.
For these two datasets, each image has an associated group attribute, such as the background of the image in the Waterbirds dataset and the gender/age of the person in the CelebA dataset.
All these attributes are correlated with the ground truth labels, but they are not directly related to the classification task.
Following previous work \cite{zhang2022contrastive}, we consider these attributes as group attributes and report the average accuracy and the worst group robustness on these datasets.
In addition, we evaluate the zero-shot domain generalization ability of our method on six datasets: CUB-200-2011 (CUB) \cite{wah2011caltech}, EuroSAT \cite{helber2019eurosat}, Place365 \cite{zhou2017places}, Flowers102 \cite{nilsback2008automated}, Food101 \cite{bossard2014food}, and Oxford Pets \cite{parkhi2012cats}.
For this task, we report the classification accuracy of the model in these datasets.

\begin{table*}[!t]
    \centering
    \caption{Average accuracy and worst group robustness on the Waterbirds and CelebA dataset.}
    \scalebox{0.85}{
    \begin{tabular}{cl|ccc|ccc|ccc|ccc}
        \toprule
       & & \multicolumn{3}{c}{RN50} & \multicolumn{3}{c}{ViT-B/32} & \multicolumn{3}{c}{ViT-B/16} & \multicolumn{3}{c}{ViT-L/14} \\ \cline{3-14}  
       & & Avg$\uparrow$   & worst$\uparrow$    & gap$\downarrow$   & Avg$\uparrow$     & worst$\uparrow$     & gap$\downarrow$    & avg$\uparrow$      & worst$\uparrow$      & gap$\downarrow$    & avg$\uparrow$      & worst$\uparrow$      & gap$\downarrow$    \\ \midrule
    \multirow{4}{*}{\rotatebox[origin=c]{90}{Waterbirds}} & CLIP &   90.47 &  16.07 & 74.40    & 87.34    & 47.28  &  40.06 &  \textbf{87.34} & 26.79 &  60.55 & 90.55 & 44.64 &  45.91 \\
    & ~~+ background  & 90.62 & 39.29 & 51.33 & 78.58   & 61.96  & \textbf{16.62} & 86.01 & 44.34 & 44.73 & 87.72 & 59.98 &  27.74 \\ \cline{3-14}
     & Ours           &   \textbf{91.76} &  44.64 & 47.30    & \textbf{89.34}    & 57.60  &   31.74 &  86.53 & 28.58 & 57.95 & \textbf{92.14} & 51.78 &  40.36 \\ 
     & ~~+ background    &   91.03 & \textbf{48.21} & \textbf{42.82}    &  85.44  & \textbf{62.50}  & 22.94 & 86.43 & \textbf{46.43} & \textbf{40.00} & 89.55 & \textbf{62.50} &  \textbf{27.05} \\ 
     \midrule
     \multirow{8}{*}{\rotatebox[origin=c]{90}{CelebA}} &CLIP &  81.05  & 73.87 & 7.18 &  80.73  & 75.82  & 4.91  & 75.16 & 62.01 & 13.15 & \textbf{86.98}  & 77.36  & 9.62  \\
     & ~~+gender & 85.97  &  81.58  & 4.39  &  80.18   & 76.18 & \textbf{4.00} & 75.92  & 66.71  &  7.99   &  80.30 & 74.31  & 5.99    \\
     & ~~+gender,age & 87.74 &  84.94 & 2.80  &  82.34 & 77.21 & 5.13 & 75.22  &  64.61  & 10.61 &  82.26 & 79.06  & 3.21    \\
     & ~~+gender,age,race & 85.91  &  82.57  & 3.34  &  81.99   & 75.67 & 6.32 & 76.37  &  67.93  &  8.44   &  82.77 & 80.00  & 2.77    \\\cline{3-14}
    & Ours & 85.35  &  83.05  & 2.30  &  \textbf{84.19}   & \textbf{78.75} & 5.44 & 79.21  &  68.54  &  10.67   &  85.79 & 81.18  & 4.61    \\
    & ~~+gender & 87.53  &  85.56  & 1.97  &  82.67   & 76.87 & 5.80 & \textbf{78.55}  & \textbf{73.84}  & \textbf{4.71}   &  81.44 & 76.14  & 5.30    \\
    & ~~+gender,age & \textbf{88.70}  & \textbf{86.35}  & 2.35  &  82.16  & 76.90 & 5.44 & 78.09  &  70.54  &  7.55   &  83.78 & 80.87  & 2.91    \\
    & ~~+gender,age,race & 85.93  &  84.18  & \textbf{1.75}  &  82.63   & 75.92 & 6.71 & 77.17  &  69.18  &  7.99   &  85.35 & \textbf{83.00}  & \textbf{2.35}    \\
    \bottomrule
    \end{tabular}}
    \label{tab:large}
\end{table*}
\noindent\textbf{Implementation Details.}
We utilize CLIP \cite{radford2021learning} as the foundation model and evaluate the performance of our method on a variety of tasks and datasets.
All experiments use PyTorch \cite{paszke2019pytorch} and are performed on a single NVIDIA A100 GPU. 
We follow the same experimental settings as the previous work \cite{an2024perceptionclip}. 
We use Resnet-50 \cite{he2016deep}, ViT-B/32, ViT-B/16, and ViT-L/14 \cite{dosovitskiy2021an} as the backbone models for evaluation of group robustness.
For the evaluation of domain generalization, we use ViT-B/16 as the backbone model. 

For the sentence embedding model, we use the Hierarchy Transformer encoder (HiT) \cite{he2024language} as the default sentence embedding model.\footnote{In our experiments, we use ``HiT-MiniLM-L12-WordNetNoun'' released on HuggingFace as the sentence embedding model.} 
Since the output of the HiT lies in the hyperbolic space, we use the logarithmic map function to transform the output to the Euclidean space \cite{yang2023hyperbolic}.
We set the dimension of the dimension of the common space in the first CCA to 64, and the dimension of the second CCA is set to the dimension of the original image embeddings.
Moreover, we set the number of random sentences $K$ to 500, and the number of random characters in each random sentence to the length of the class name (similar to that of \cite{roth2023waffling,an2024perceptionclip}).

\subsection{Results on Group Robustness}\label{sec:result-1}
We first evaluate the group robustness of our method on the Waterbirds and CelebA datasets.
The results are reported in Table \ref{tab:large}.
We mainly evaluate four different backbone models (RN50, ViT-B/32, ViT-B/16, and ViT-L/14). 
The results are compared between the baseline CLIP model and our proposed method (DoubleCCA).

First, we show the results when the text prompts only describe the class and ignore the contextual attributes. 
First, we observe that our method achieves better average accuracy and worst group robustness than the baseline CLIP model on both datasets.
Although the average accuracy of our method is slightly lower than that of the baseline CLIP model, when the backbone is ViT-B/16 on Waterbirds and the backbone is ViT-L/14 on CelebA, the worst group robustness is significantly improved.
We think this is a trade-off between average accuracy and worst group robustness, which has also been observed in recent work \cite{dehdashtian2024utility}. 
For example, when the backbone is ViT-L/14 on CelebA, the worst group robustness of our method is 81.18\%, which is higher than that of the baseline CLIP model (77.36\%). 
However, the average accuracy has a slight decrease (from 86.98\% to 85.79\%) compared to the baseline CLIP model.  

Following PerceptionCLIP \cite{an2024perceptionclip}, we include contextual attributes such as conditional information, such as background information in the Waterbirds dataset and gender information (i.e., female and male) in the CelebA dataset. 
Here, we only substitute the original prompt embedding with the merged text embeddings $W$ in the CLIP model and then use the same inference process as in \cite{an2024perceptionclip}.

We report the results on Waterbirds by considering the background as the contextual attributes, such as in forest, in sky, on street, on grass, on tree, with folowers, on beach, with human, on a branch, etc.
First, the same phenomena are observed in our reproduced results, where the group robustness can be improved by incorporating these attributes, which also help reduce the accuracy gap and achieve a more fair zero-shot classifier.
Second, we observe that by using our method, the worst group robustness can be further improved, and the accuracy gap can be further reduced in most cases.
More interestingly, considering the backgroup information, the worst group robustness has a consistent improvement across different backbone models, but the average accuracy has a slight decrease. 
Thus, in this case, a trade-off between average accuracy and worst group robustness is also observed. But we think this will be of benefit to achieve a more fair zero-shot classifier.

Then, we also report the results on CelebA by considering contextual attributes, such as gender (female and male), age (young and old), race (white skin, dark skin, Asian, and others), etc.
We observe that our method can achieve overall better average accuracy and worst group robustness than the baseline CLIP model on the CelebA dataset.
For instance, when the backbone is ViT-B/16, the accuracy gap of our method is 4.71\%, which is lower than that of the baseline CLIP model (7.99\%), considering the contextual attribute of gender. 
Furthermore, compared to the results shown in FairerCLIP, our method achieves better results when the backbone is ResNet-50, where the best worst group robustness of our method is 86.35\%, which is higher than that of FairerCLIP (81.50\%). 
For the backbone of ViT-L/14, our method also achieves a competitive result compared to FairerCLIP, where the best worst group robustness of our method is 83. 00\%, which is slightly lower than that of FairerCLIP (85.20\%).
It is worth noting that FairerCLIP utilizes the target label and attributes to learn a kernel map function in supervised way, which is more complex than our method.

\begin{figure}[!t]
    \centering
    \includegraphics[width=0.47\textwidth]{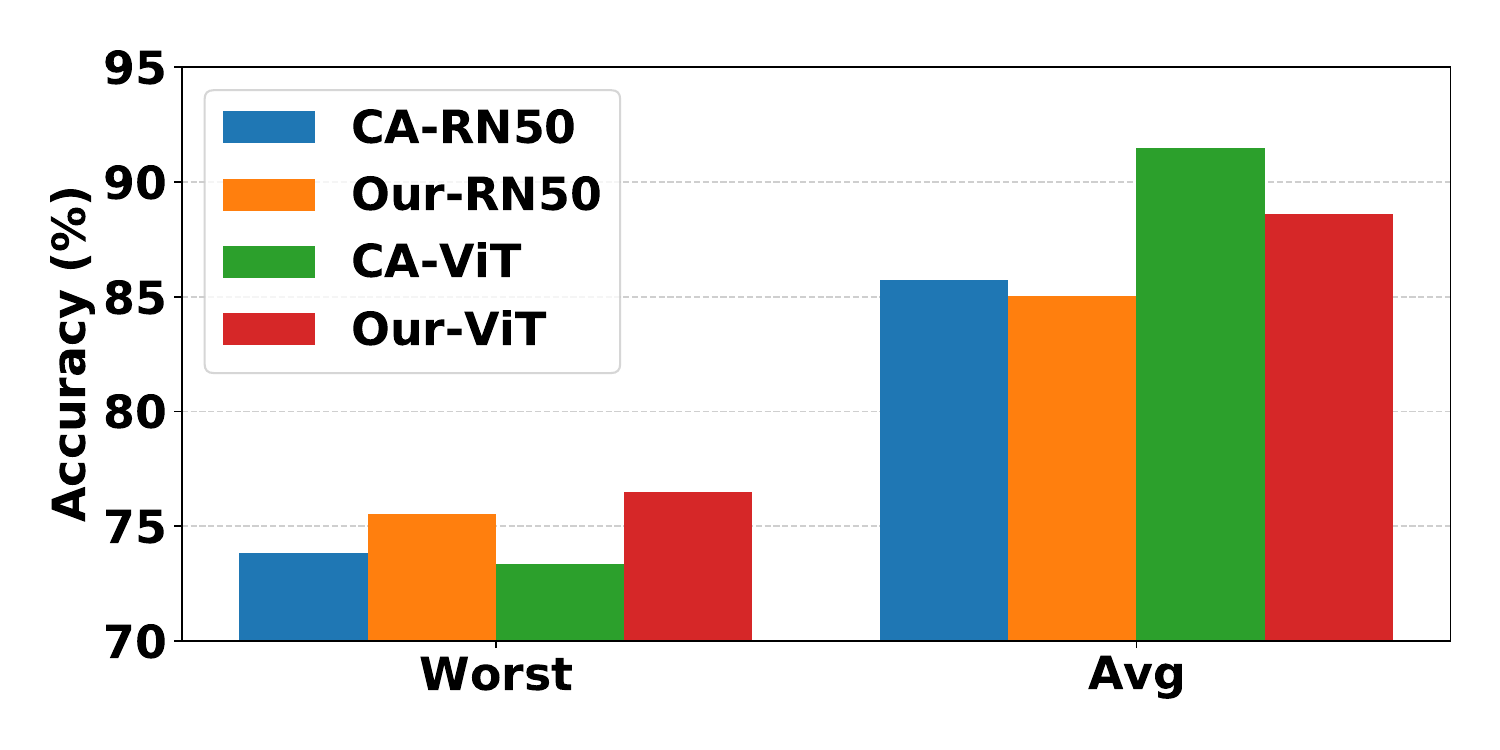}
    \vspace{-1em}
    \caption{Combination of Contrastive Adapter (CA) and our proposed DoubleCCA. We report the average accuracy and worst group robustness on the Waterbirds dataset. The backbone model is ViT-L/14 and ResNet-50.
    }
    \label{fig:CA}
\end{figure}
Since our method can be easily integrated into existing models, we also combine our method with the contrastive adapter (CA) \cite{zhang2022contrastive} to further improve the group robustness of the CLIP model.
In detail, we first use DoubleCCA to generate the merged text embeddings, and then substitute the original text embeddings with the merged text embeddings in the CLIP model. Finally, we use the CA algorithm to learn the adapter. 
The results are shown in Figure \ref{fig:CA}.
We observe that using the merged text embeddings helps improve the worst group accuracy, but the average accuracy has a slight decrease.
Thus, in this case, the trade-off between the average accuracy and the worst group robustness is also observed.

Overall, the results demonstrate that DoubleCCA effectively enhances the group robustness of foundation models, providing better performance and fairness across different datasets and backbone models.
Moreover, in different scenarios, trade-off phenomena are observed, which is consistent with previous work \cite{dehdashtian2024utility}.

\begin{table}[!t]
    \centering
    \caption{Classification accuracy of ViT-B/16 on different data domains with foundation models.}
    \begin{tabular}{c|cc}
        \toprule
            & CLIP & Ours \\ \hline
            CUB & 56.67 & 56.30 \\ 
            EuroSAT & 51.44 & 52.84 \\ 
            Place365 & 38.93 & 39.47 \\ 
            Flowers102 & 67.73 & 68.82 \\ 
            Food101 & 88.24 & 88.35 \\ 
            Oxford Pets & 88.25 & 88.57 \\ \bottomrule
    \end{tabular}
    \label{tab:general}
\end{table}
\subsection{Results on Domain Generalization}\label{sec:result-2}
We evaluate the domain generalization ability of our method on six widely used datasets, i.e., CUB-200-2011 (CUB), EuroSAT, Place365, Flowers102, Food101, and Oxford Pets.

We report the classification accuracy of the ViT-B/16 backbone model in these data sets in Table \ref{tab:general}.
We observe that our method achieves competitive results compared to the baseline CLIP model on these datasets.
For example, the classification accuracy of our method is 56. 30\% in the CUB dataset, which is slightly lower than that of the baseline CLIP model (56.67\%).
Furthermore, our method achieves better results on the EuroSAT, Place365, Flowers102, Food101, and Oxford Pets datasets, where the classification accuracy of our method is 52.84\%, 39.47\%, 68.82\%, 88.35\%, and 88.57\%, respectively.

By comparing and analyzing Tables 1 and 2, we can summarize the following.
Our method maintains the generalizability of the foundation model while improving performance and group robustness across various datasets and backbone architectures.

\subsection{Effect of Sentence Embeddings}\label{sec:analysis}
Since DoubleCCA leverages sentence embeddings to enhance the text embeddings of the CLIP model, we conduct an ablation study to analyze the effect of sentence embeddings on the group robustness of the CLIP model.

\begin{figure}[!t]
    \centering
    \includegraphics[width=0.4\textwidth]{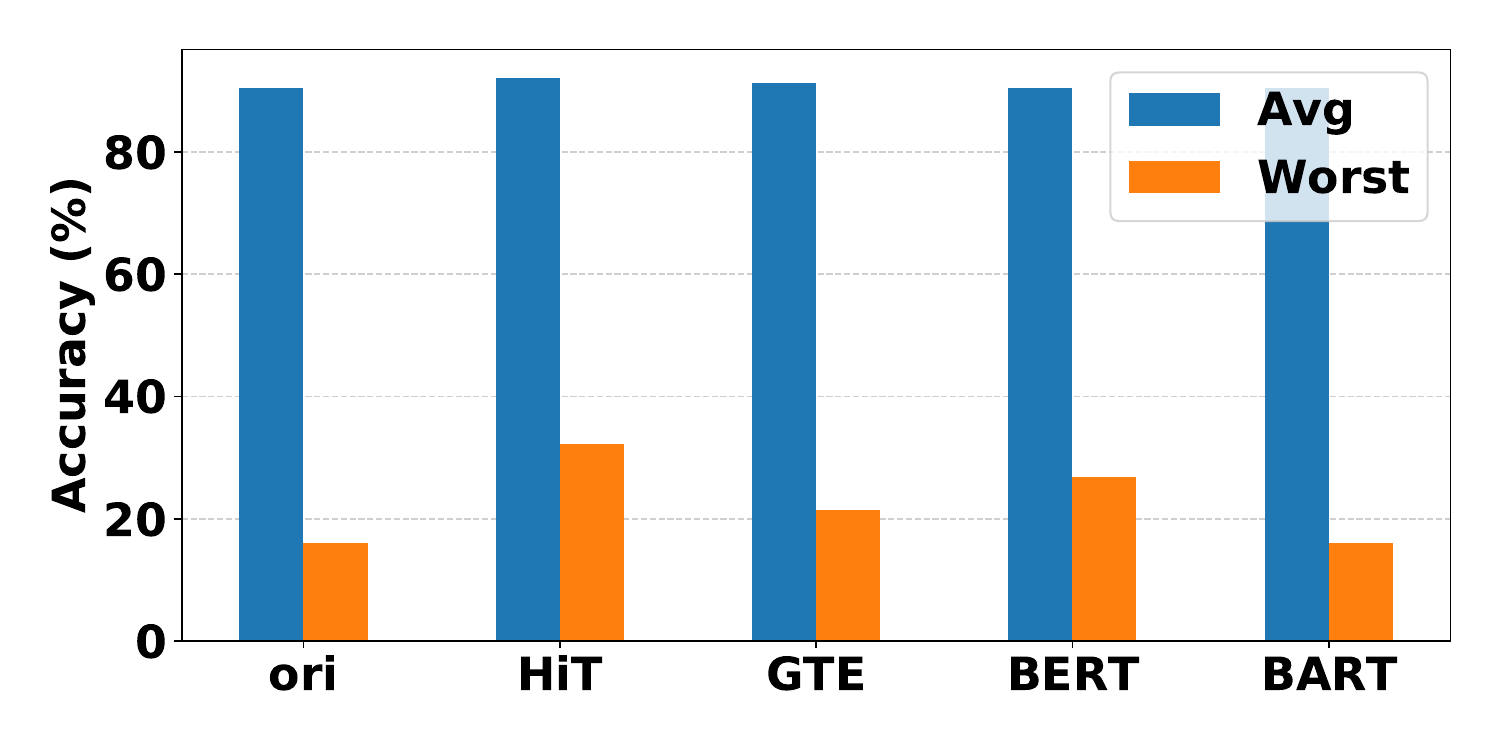}
    \vspace{-1.em}
    \caption{Ablation study results on the Waterbirds dataset.}
    \label{fig:ablation_1}
\end{figure}

In previous experiments, we use the HiT model \cite{he2024language} to generate sentence embeddings. 
To further study the effect of sentence embeddings, we replace the HiT model with other sentence embedding models. 
To ensure a comprehensive comparison, we select popular models from HuggingFace Hub\footnote{\url{https://huggingface.co/}} as alternatives to the default HiT model, such as the classical Sentence-BERT model \cite{sentence-bert}, gte-base-en-v1.5 model \cite{li2023towards}, and bart-base model \cite{bart}.
We directly use the pre-trained models released by HuggingFace Hub to generate sentence embeddings for the Waterbirds dataset.
The results are shown in Figure \ref{fig:ablation_1}.

Compared with the original CLIP model, we observe that different sentence embedding methods in DoubleCCA either improve the model's performance or maintain it at a comparable level.
Notably, HiT demonstrate the most significant improvements in performance. Both Sentence-BERT and gte-base-en-v1.5 also have a positive impact on the model's performance. 

First, HiT is a state-of-the-art sentence embedding model that aims to learn the hierarchical semantic structure in language models. 
HiT is trained on WordNet, which can provide unseen subsumptions and hypernyms for the words in the sentence. 
Second, Sentence-BERT and gte-base models are also popular sentence embedding models, which are verified to be effective in unsupervised text retrieval tasks.
However, BART shows little improvement in model performance. We think this is because BART targets dialogue understanding, question answering, and summarization tasks, which may face the same problems as mentioned before, where it will introduce semantic ambiguity to text embeddings \cite{menon2022visual}.

Overall, the results demonstrate that the choice of the sentence embedding model can significantly affect the performance of the foundation model.
We recommend using HiT as the default sentence embedding model in DoubleCCA, as it achieves the best performance in our experiments.
Moreover, it is more interesting to explore the effect of different sentence embedding models on the group robustness of the foundation model, which is left for future work.

\subsection{Abalation Study}\label{sec:ablation}

\begin{figure}
    \centering
    \begin{tabular}{ccc}
        \includegraphics[width=0.2\textwidth]{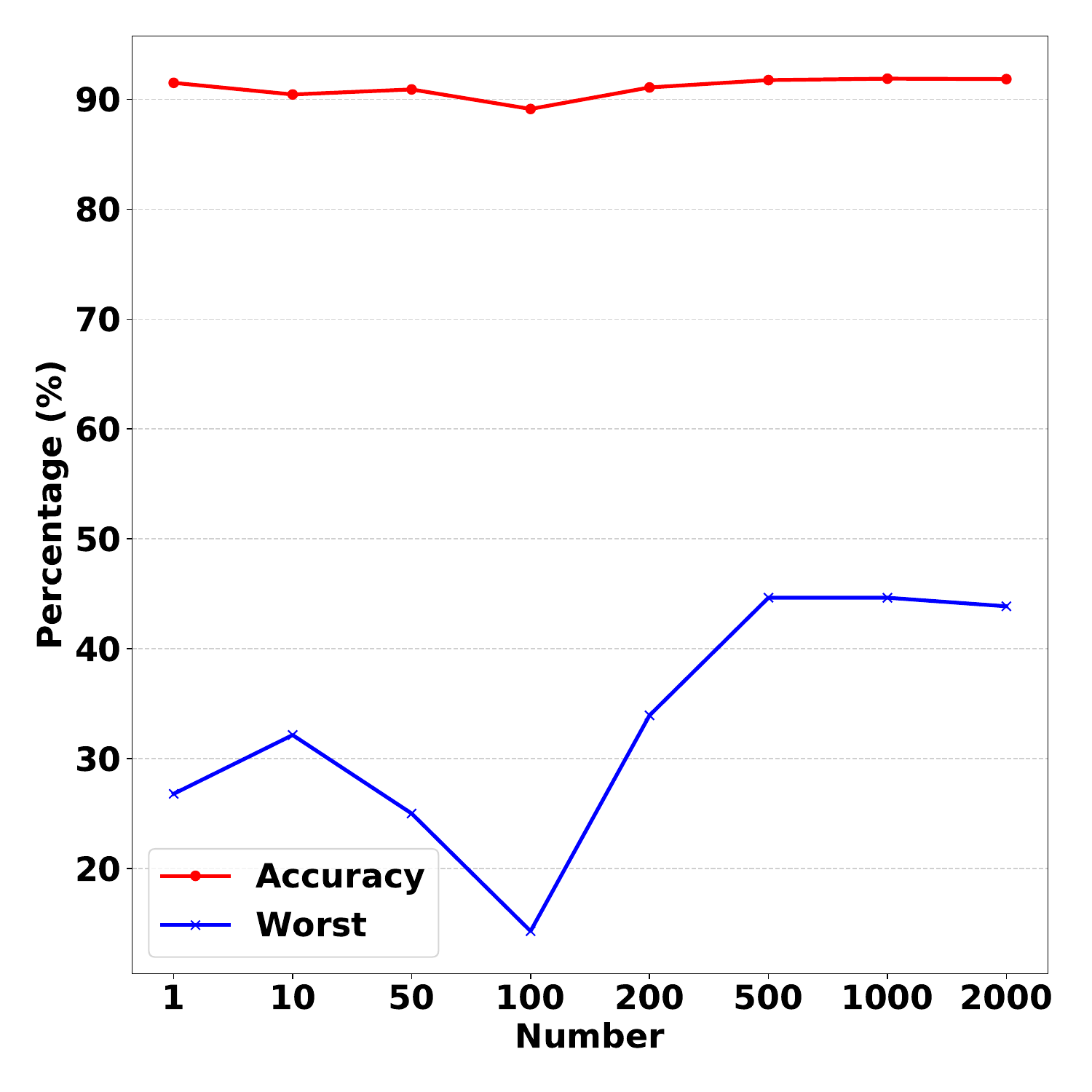} & \includegraphics[width=0.2\textwidth]{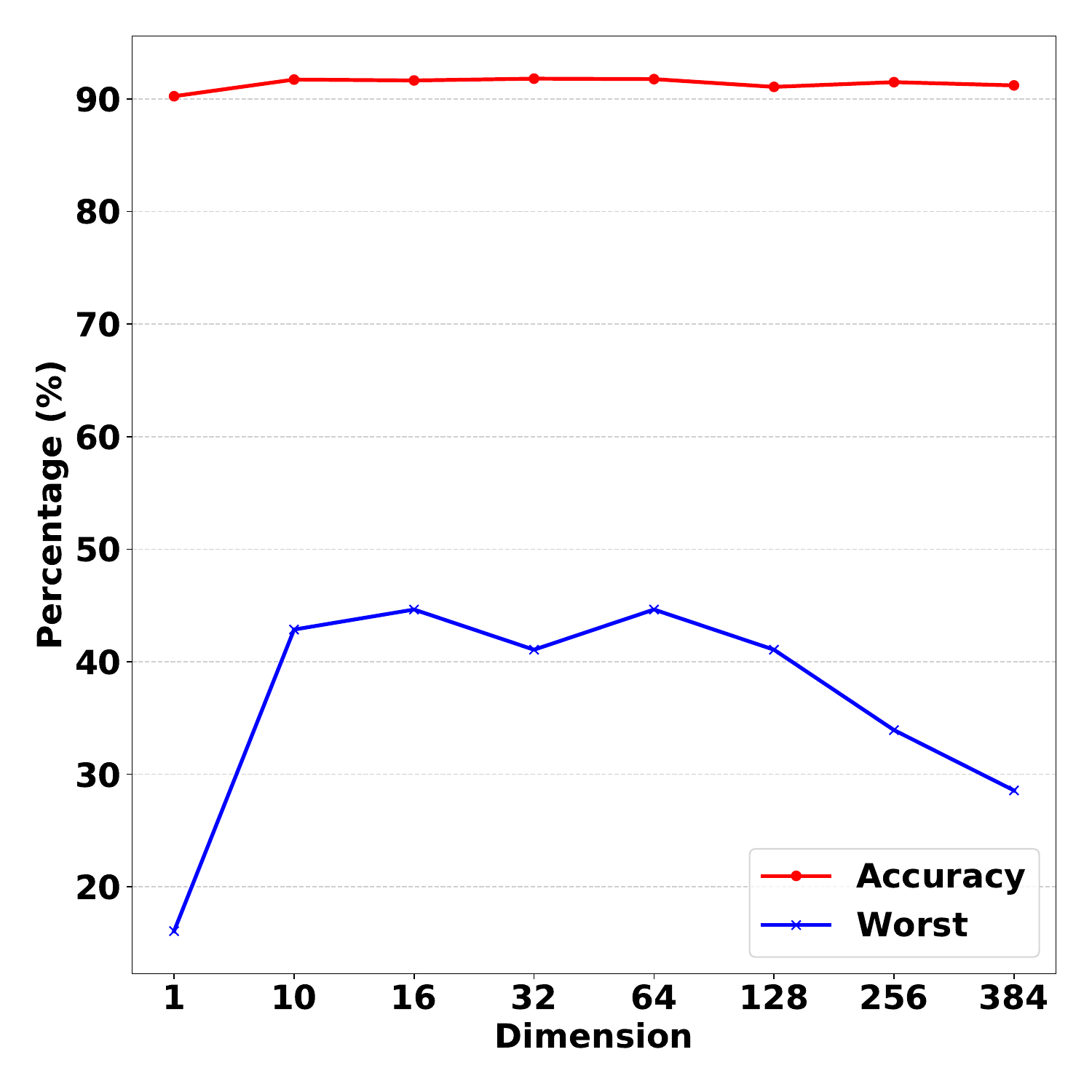} & \includegraphics[width=0.43\textwidth]{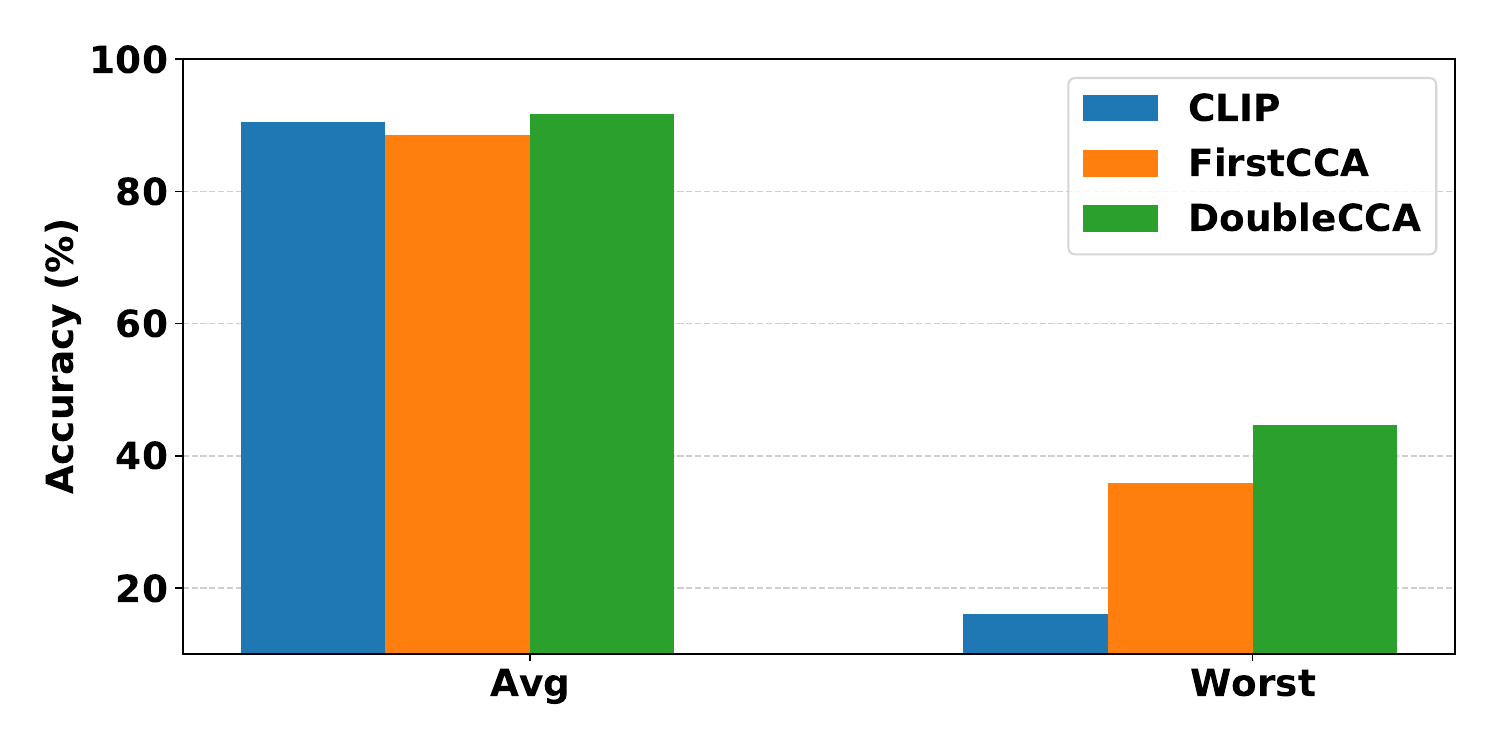}\\
        (a) \# of random sentences & (b) Dimension of CCA & (c) Effect of the first CCA\\
    \end{tabular}  
    \caption{Ablation study results on the Waterbirds dataset. \label{fig:ab2}} 
\end{figure}

\subsubsection{Effect of the Hyperparameters.}

\noindent\textbf{Number of Random Sentences.}
We conduct an ablation study to analyze the effect of the number of random sentences on the group robustness. 
We employ the backbone model for ResNet-50 and fix the dimension of the CCA as 64.
Then, we vary the number of random sentences from 1 to 2000.
The results are shown in Figure \ref{fig:ab2} (a).

The results indicate that varying the number of random sentences has minimal impact on the average accuracy but demonstrates a substantial influence on the worst group robustness.
When the number of random sentences is less than 500, the worst group robustness exhibits high variability. 
In particular, when the number of sentences drops to 100, the performance deteriorates below that of the original CLIP model.
We attribute this instability to the inherent randomness of random character sequences. However, as the number of sentences increases, the model performance gradually stabilizes, suggesting that sufficient random sentences enable the model to capture meaningful pragmatic information.

\noindent\textbf{Dimension of CCA.}
We further study the effect of the dimension of the CCA on the group robustness of the CLIP model.
We employ the ResNet-50 backbone model and fix the number of random sentences to 500.
Then, we vary the dimension of the CCA from 1 to 384\footnote{The dimension of the HiT feature is 384.}.
The results are shown in Figure \ref{fig:ab2} (b).
The results indicate that the dimension of the common space significantly impacts performance. 
Both low and high dimensions adversely affect the results; low dimensions lead to insufficient feature representation, while high dimensions introduce feature vectors corresponding to small singular values. 
We recommend setting the dimension of the CCA to 64, as it achieves the best performance in our experiments. Moreover, as discussed in \cite{vidal2016principal},
the dimension of this subspace is a natural measure of the model complexity, thus some automatic dimension selection methods can be used to determine the optimal dimension of the CCA. We leave this for future work. 

\subsubsection{Effect of the First CCA.}
Finally, we analyze the effect of the first CCA on the group robustness of the CLIP model.
We employ the backbone model to ResNet-50 and fix the number of random sentences to 500.
Then, we remove the second CCA from the DoubleCCA method and directly use Eq.\ref{eq3} as the score function for the zero-shot classification.
The results are shown in Figure \ref{fig:ab2} (c).
The results indicate that only the first CCA also has a positive impact on the group robustness of the CLIP model.
But the second CCA step is essential for further improving the group robustness of the CLIP model.

\section{Related Work}\label{sec:related}
This section will briefly review related work in group robustness. 
Group robustness is a critical issue in machine learning, especially in the context of fairness and bias.
There are many works that focus on improving the group robustness of foundation models.
Existing method can be divided into two categories: prompt tuning, adapter-based methods, and fine-tuning methods.
The first category includes methods that modify the input prompts given to a pre-trained model to better align with the desired output. Representative works include \cite{chuang2023debiasing,pmlr-v235-phan24b,yang2024debiasing}.
The second category includes methods that add additional modules to the pre-trained model to adapt it to the target task. Representative works include \cite{zhang2022contrastive,gao2024clip,dehdashtian2024fairerclip}.
The third category includes methods that fine-tune the pre-trained model on the target task. The representative works include \cite{kumar2022finetuning}.
In addition to these methods, An et al. \cite{an2024perceptionclip} proposes a perception-aware method (called PerceptionCLIP) to enhance the group robustness of the CLIP model, which provides CLIP with contextual attributes. 
This is similar to our method, which also enriches the text embeddings of the CLIP model with additional semantic information. Both of us aim to improve the group robustness of the zero-shot classifier. According to the experiments, our method outperforms PerceptionCLIP in terms of both average accuracy and worst group robustness.
Since our method is simple and easy to implement, it can be easily integrated into existing models, such as the contrastive adapter \cite{zhang2022contrastive}, providing a practical solution to improve the robustness of the foundation models.

\section{Conclusion}\label{sec:conclusion}
In this paper, we proposed DoubleCCA, a novel method to improve the robustness of foundation models to group-based biases. By leveraging random sentence embeddings and employing Canonical Correlation Analysis (CCA) twice, our method effectively aligns and merges different text representations. We demonstrated the effectiveness of DoubleCCA on various datasets, showing that it outperforms existing methods in terms of both group robustness and domain generalization. Our approach is simple to implement and can be easily integrated into existing models, providing a practical solution to improve the robustness of foundation models.

\bibliographystyle{unsrt}
\bibliography{reference}

\begin{thebibliography}{10}

\bibitem{radford2021learning}
Alec Radford, Jong~Wook Kim, Chris Hallacy, Aditya Ramesh, Gabriel Goh, Sandhini Agarwal, Girish Sastry, Amanda Askell, Pamela Mishkin, Jack Clark, et~al.
\newblock Learning transferable visual models from natural language supervision.
\newblock In {\em ICML}, 2021.

\bibitem{zhai2023sigmoid}
Xiaohua Zhai, Basil Mustafa, Alexander Kolesnikov, and Lucas Beyer.
\newblock Sigmoid loss for language image pre-training.
\newblock In {\em ICCV}, 2023.

\bibitem{desai2023hyperbolic}
Karan Desai, Maximilian Nickel, Tanmay Rajpurohit, Justin Johnson, and Shanmukha~Ramakrishna Vedantam.
\newblock Hyperbolic image-text representations.
\newblock In {\em ICML}, 2023.

\bibitem{schuhmann2021laion}
Christoph Schuhmann, Richard Vencu, Romain Beaumont, Robert Kaczmarczyk, Clayton Mullis, Aarush Katta, Theo Coombes, Jenia Jitsev, and Aran Komatsuzaki.
\newblock Laion-400m: Open dataset of clip-filtered 400 million image-text pairs.
\newblock {\em arXiv:2111.02114}, 2021.

\bibitem{wang2022medclip}
Zifeng Wang, Zhenbang Wu, Dinesh Agarwal, and Jimeng Sun.
\newblock Medclip: Contrastive learning from unpaired medical images and text.
\newblock {\em arXiv:2210.10163}, 2022.

\bibitem{ramaswamy2024geode}
Vikram~V Ramaswamy, Sing~Yu Lin, Dora Zhao, Aaron Adcock, Laurens van~der Maaten, Deepti Ghadiyaram, and Olga Russakovsky.
\newblock Geode: a geographically diverse evaluation dataset for object recognition.
\newblock {\em NeurIPS}, 2024.

\bibitem{sun2024clip}
Shuyang Sun, Runjia Li, Philip Torr, Xiuye Gu, and Siyang Li.
\newblock Clip as rnn: Segment countless visual concepts without training endeavor.
\newblock In {\em CVPR}, 2024.

\bibitem{licascade}
Yunheng Li, Zhong-Yu Li, Quan-Sheng Zeng, Qibin Hou, and Ming-Ming Cheng.
\newblock Cascade-clip: Cascaded vision-language embeddings alignment for zero-shot semantic segmentation.
\newblock In {\em ICML}, 2024.

\bibitem{menon2022visual}
Sachit Menon and Carl Vondrick.
\newblock Visual classification via description from large language models.
\newblock In {\em ICLR}, 2022.

\bibitem{roth2023waffling}
Karsten Roth, Jae~Myung Kim, A~Koepke, Oriol Vinyals, Cordelia Schmid, and Zeynep Akata.
\newblock Waffling around for performance: Visual classification with random words and broad concepts.
\newblock In {\em ICCV}, 2023.

\bibitem{an2024perceptionclip}
Bang An, Sicheng Zhu, Michael-Andrei Panaitescu-Liess, Chaithanya~Kumar Mummadi, and Furong Huang.
\newblock Perception{CLIP}: Visual classification by inferring and conditioning on contexts.
\newblock In {\em ICLR}, 2024.

\bibitem{zhang2022contrastive}
Michael Zhang and Christopher R{\'e}.
\newblock Contrastive adapters for foundation model group robustness.
\newblock {\em NeurIPS}, 2022.

\bibitem{kumar2022finetuning}
Ananya Kumar, Aditi Raghunathan, Robbie~Matthew Jones, Tengyu Ma, and Percy Liang.
\newblock Fine-tuning can distort pretrained features and underperform out-of-distribution.
\newblock In {\em ICLR}, 2022.

\bibitem{kirichenko2023last}
Polina Kirichenko, Pavel Izmailov, and Andrew~Gordon Wilson.
\newblock Last layer re-training is sufficient for robustness to spurious correlations.
\newblock In {\em ICLR}, 2023.

\bibitem{chuang2023debiasing}
Ching-Yao Chuang, Varun Jampani, Yuanzhen Li, Antonio Torralba, and Stefanie Jegelka.
\newblock Debiasing vision-language models via biased prompts.
\newblock {\em arXiv:2302.00070}, 2023.

\bibitem{dehdashtian2024fairerclip}
Sepehr Dehdashtian, Lan Wang, and Vishnu Boddeti.
\newblock Fairer{CLIP}: Debiasing {CLIP}'s zero-shot predictions using functions in {RKHS}s.
\newblock In {\em ICLR}, 2024.

\bibitem{you2024calibrating}
Chenyu You, Yifei Mint, Weicheng Dai, Jasjeet~S Sekhon, Lawrence Staib, and James~S Duncan.
\newblock Calibrating multi-modal representations: A pursuit of group robustness without annotations.
\newblock In {\em CVPR}, 2024.

\bibitem{gao2024clip}
Peng Gao, Shijie Geng, Renrui Zhang, Teli Ma, Rongyao Fang, Yongfeng Zhang, Hongsheng Li, and Yu~Qiao.
\newblock Clip-adapter: Better vision-language models with feature adapters.
\newblock {\em IJCV}, 2024.

\bibitem{pmlr-v235-phan24b}
Hoang Phan, Andrew~Gordon Wilson, and Qi~Lei.
\newblock Controllable prompt tuning for balancing group distributional robustness.
\newblock In {\em ICML}, 2024.

\bibitem{yang2024debiasing}
Yunfan Yang, Chaoquan Jiang, Zhiyu Lin, Jinlin Xiao, Jiaming Zhang, and Jitao Sang.
\newblock Debiasing vison-language models with text-only training.
\newblock {\em arXiv:2410.09365}, 2024.

\bibitem{he2024language}
Yuan He, Zhangdie Yuan, Jiaoyan Chen, and Ian Horrocks.
\newblock Language models as hierarchy encoders.
\newblock In {\em NeurIPS}, 2024.

\bibitem{corrochano2005eigenproblems}
Eduardo~Bayro Corrochano, Tijl De~Bie, Nello Cristianini, and Roman Rosipal.
\newblock Eigenproblems in pattern recognition.
\newblock {\em Handbook of Geometric Computing: Applications in Pattern Recognition, Computer Vision, Neuralcomputing, and Robotics}, 2005.

\bibitem{horoi2024harmony}
Stefan Horoi, Albert Manuel~Orozco Camacho, Eugene Belilovsky, and Guy Wolf.
\newblock Harmony in diversity: Merging neural networks with canonical correlation analysis.
\newblock In {\em ICML}, 2024.

\bibitem{Sagawa2020Distributionally}
Shiori Sagawa, Pang~Wei Koh, Tatsunori~B. Hashimoto, and Percy Liang.
\newblock Distributionally robust neural networks.
\newblock In {\em ICLR}, 2020.

\bibitem{maaten2008visualizing}
Laurens van~der Maaten and Geoffrey Hinton.
\newblock Visualizing data using t-sne.
\newblock {\em JMLR}, 2008.

\bibitem{sentence-bert}
Nils Reimers and Iryna Gurevych.
\newblock Making monolingual sentence embeddings multilingual using knowledge distillation.
\newblock In {\em EMNLP}, 2020.

\bibitem{liu2015deep}
Ziwei Liu, Ping Luo, Xiaogang Wang, and Xiaoou Tang.
\newblock Deep learning face attributes in the wild.
\newblock In {\em ICCV}, 2015.

\bibitem{wah2011caltech}
Catherine Wah, Steve Branson, Peter Welinder, Pietro Perona, and Serge Belongie.
\newblock The caltech-ucsd birds-200-2011 dataset.
\newblock 2011.

\bibitem{helber2019eurosat}
Patrick Helber, Benjamin Bischke, Andreas Dengel, and Damian Borth.
\newblock Eurosat: A novel dataset and deep learning benchmark for land use and land cover classification.
\newblock {\em IEEE J. Sel. Top. Appl. Earth Obs. Remote Sens}, 2019.

\bibitem{zhou2017places}
Bolei Zhou, Agata Lapedriza, Aditya Khosla, Aude Oliva, and Antonio Torralba.
\newblock Places: A 10 million image database for scene recognition.
\newblock {\em IEEE TPAMI}, 2017.

\bibitem{nilsback2008automated}
Maria-Elena Nilsback and Andrew Zisserman.
\newblock Automated flower classification over a large number of classes.
\newblock In {\em ICVGIP}, 2008.

\bibitem{bossard2014food}
Lukas Bossard, Matthieu Guillaumin, and Luc Van~Gool.
\newblock Food-101--mining discriminative components with random forests.
\newblock In {\em ECCV}, 2014.

\bibitem{parkhi2012cats}
Omkar~M Parkhi, Andrea Vedaldi, Andrew Zisserman, and CV~Jawahar.
\newblock Cats and dogs.
\newblock In {\em CVPR}, 2012.

\bibitem{paszke2019pytorch}
Adam Paszke, Sam Gross, Francisco Massa, Adam Lerer, James Bradbury, Gregory Chanan, Trevor Killeen, Zeming Lin, Natalia Gimelshein, Luca Antiga, et~al.
\newblock Pytorch: An imperative style, high-performance deep learning library.
\newblock {\em NeurIPS}, 2019.

\bibitem{he2016deep}
Kaiming He, Xiangyu Zhang, Shaoqing Ren, and Jian Sun.
\newblock Deep residual learning for image recognition.
\newblock In {\em CVPR}, 2016.

\bibitem{dosovitskiy2021an}
Alexey Dosovitskiy, Lucas Beyer, Alexander Kolesnikov, Dirk Weissenborn, Xiaohua Zhai, Thomas Unterthiner, Mostafa Dehghani, Matthias Minderer, Georg Heigold, Sylvain Gelly, Jakob Uszkoreit, and Neil Houlsby.
\newblock An image is worth 16x16 words: Transformers for image recognition at scale.
\newblock In {\em ICLR}, 2021.

\bibitem{yang2023hyperbolic}
Menglin Yang, Min Zhou, Rex Ying, Yankai Chen, and Irwin King.
\newblock Hyperbolic representation learning: Revisiting and advancing.
\newblock In {\em ICML}, 2023.

\bibitem{dehdashtian2024utility}
Sepehr Dehdashtian, Bashir Sadeghi, and Vishnu~Naresh Boddeti.
\newblock Utility-fairness trade-offs and how to find them.
\newblock In {\em CVPR}, 2024.

\bibitem{li2023towards}
Zehan Li, Xin Zhang, Yanzhao Zhang, Dingkun Long, Pengjun Xie, and Meishan Zhang.
\newblock Towards general text embeddings with multi-stage contrastive learning.
\newblock {\em arXiv:2308.03281}, 2023.

\bibitem{bart}
Mike Lewis, Yinhan Liu, Naman Goyal, Marjan Ghazvininejad, Abdelrahman Mohamed, Omer Levy, Veselin Stoyanov, and Luke Zettlemoyer.
\newblock {BART}: Denoising sequence-to-sequence pre-training for natural language generation, translation, and comprehension.
\newblock In {\em ACL}, 2020.

\bibitem{vidal2016principal}
Ren{\'e} Vidal, Yi~Ma, S~Shankar Sastry, Ren{\'e} Vidal, Yi~Ma, and S~Shankar Sastry.
\newblock Principal component analysis.
\newblock {\em Generalized principal component analysis}, 2016.

\end{thebibliography}


\end{document}